\documentclass[10pt, a4paper]{article}
\usepackage{lscape}

\usepackage[final]{lrec2026} % this is the new style
\usepackage{xcolor}

\newcommand{\ag}{\textcolor{violet}{Ag}}
\newcommand{\pat}{\textcolor{orange}{Pat}}
\newcommand{\akt}{\textcolor{brown}{Akt}}
\newcommand{\pass}{\textcolor{teal}{Pass}}
\newcommand{\prep}{\textcolor{blue}{p}}
\newcommand{\by}{\textcolor{red}{by}}

\title{Datasets for Verb Alternations across Languages:\\ BLM Templates and Data Augmentation Strategies}

\name{Giuseppe Samo$^{1}$, Paola Merlo$^{1,2}$} 

\address{$^{1}$IDIAP Research Institute, $^{2}$University of Geneva \\
          \{giuseppe.samo, paola.merlo\}@idiap.ch\\}

\abstract{
Large language models (LLMs) have shown remarkable performance across various sentence-based linguistic phenomena, yet their ability to capture cross-sentence paradigmatic patterns, such as verb alternations,  remains underexplored. In this work, we present curated paradigm-based datasets for four languages, designed to probe systematic cross-sentence knowledge of  verb alternations (change-of-state and object-drop constructions in English, German  and Italian, and Hebrew binyanim). The datasets comprise thousands of the Blackbird Language Matrices (BLMs) problems.  The BLM task -- an RPM/ARC-like task devised specifically for language -- is a controlled linguistic puzzle where models must select the sentence that completes a pattern according to syntactic and semantic rules. We introduce three types of templates varying in complexity and apply linguistically-informed data augmentation strategies across synthetic and natural data. We provide simple baseline performance results across English, Italian, German, and Hebrew, that demonstrate the diagnostic usefulness of the datasets.
 \\ \newline \Keywords{Verb Alternation, Synthetic curated datasets, Corpus Data, LLMs, Syntax/Semantics} }

\begin{document}

\maketitleabstract

\section{Introduction}
Large language models (LLMs)  have demonstrated successful performance in a vast range of practical language-based applications and the nature and extent of their linguistic competence have been investigated across a wide range of linguistic phenomena, with impressive results in areas such as negation, subject-verb agreement or other dependency-related phenomena \citep{warstadt-etal-2019-investigating, ravichander-etal-2021-probing, belinkov-2022-probing, Acs_Hamerlik_Schwartz_Smith_Kornai_2024,linzenbaroni21, morante2021recent, gautam2024subject}.  

A further dimension of complexity arises with paradigmatic phenomena such as verb alternations or inflectional patterns, since uncovering the underlying rules requires considering relations across multiple sentences rather than within a single sentence. These cross-sentence dependencies reveal broader systematic structures that go beyond local grammatical constraints \citep{bobaljik2015suppletion,klein-tsarfaty-2020-getting}. In this work, we focus on causative alternations, a phenomenon where certain verbs can appear in different forms and structures, one of them conveying causative meaning. Consider the English verbs \textit{paint} and \textit{break}, which can be transitive or intransitive. In the transitive use, an agent acts on a patient, causing a change of state (e.g., \textit{The artist painted/broke the vase}). In the intransitive use, \textit{paint} retains its agentive subject (\textit{The artist painted}), but \textit{break} shows the patient surfacing as the subject of the clause (\textit{The vase broke}). This alternation in \textit{break}, a change-of-state verb (COS), involves an argument shift, that does not occur for verbs  like \textit{paint} (Object Drop class, OD) \cite{Levin93, merlo2001automatic}.

%Consider, for instance, the English verbs \textit{paint} and \textit{break} that can appear in both transitive and intransitive forms. In its transitive use, an agent acts upon a patient, typically causing a change of state (e.g., \textit{The artist\textsubscript{\textcolor{violet}{Agent}} painted the vase\textsubscript{\textcolor{orange}{Patient}}}, \textit{The artist\textsubscript{\textcolor{violet}{Agent}} broke the  vase\textsubscript{\textcolor{orange}{Patient}}}). In the intransitive use, while the verb \textit{paint} shows the same agentive subject (\textit{The artist\textsubscript{\textcolor{violet}{Agent}} painted}), for the verb \textit{break} the patient surfaces as the syntactic subject (\textit{The vase\textsubscript{\textcolor{orange}{Patient}} broke}). The alternation of the verb \textit{break}-- a change of state verb \citep{Levin93} -- involves an argument shift between the two configurations: the external argument is suppressed, and the internal argument is promoted to subject position. By contrast, no such shift occurs in other verb classes allowing an alternation between transitive and intransitive forms like the one of the verb \textit{paint} -- Object Drop class \citep{Levin93,merlo2001automatic} -- where both forms share the same agentive subject.

\begin{figure*}[h!]
    \centering
    \begin{tabular}{|l|l|l|l|}
    \hline
    \textbf{Language} & \textbf{Structure} & \textbf{Change of state verbs} & \textbf{Object Drop verbs} \\ \hline
    English & Transitive & The \textcolor{violet}{artist} \textbf{broke} the \textcolor{orange}{vase} & The \textcolor{violet}{artist} \textbf{paints} the \textcolor{orange}{vase} \\
    & Intransitive & The \textcolor{orange}{vase} \textbf{broke} & The \textcolor{violet}{artist} \textbf{paints} \\ \hline
    Italian & Transitive & L' \textcolor{violet}{artista} \textbf{ruppe} il \textcolor{orange}{vaso} & L' \textcolor{violet}{artista} \textbf{dipingeva} il \textcolor{orange}{vaso}  \\
    & Intransitive & Il \textcolor{orange}{vaso} \textbf{\textcolor{red}{si} ruppe} & L' \textcolor{orange}{artista} \textbf{dipingeva} \\ \hline
    German & Transitive & \textcolor{blue}{Der} \textcolor{violet}{Künstler} \textbf{zerbrach} \textcolor{teal}{den} \textcolor{orange}{Becher} & \textcolor{blue}{Der} \textcolor{violet}{Künstler} \textbf{malte} \textcolor{teal}{den} \textcolor{orange}{Becher}\\
    & Intransitive & \textcolor{blue}{Der} \textcolor{orange}{Becher} \textbf{zerbrach} & \textcolor{blue}{Der} \textcolor{violet}{Künstler} \textbf{malte} \\ \hline
    \end{tabular}
    \caption{Example of Change of state verbs and Object drop verbs alternation and morphological marking across languages. The verb is highlighted in bold while colours code semantic roles (\textcolor{violet}{agents}, \textcolor{orange}{patients}), morphological marking on the verb (\textcolor{red}{si} in Italian) and morphological case marking on the argument \textcolor{blue}{nominative} and \textcolor{teal}{accusative} case.}
    \label{fig:intro-example}
\end{figure*}

Languages differ considerably in how they encode the causative alternation morphologically  \citep{haspelmath2014coding, samardvzic2018probability}. Figure~\ref{fig:intro-example} illustrates this variation in causative and object drop alternations across English, Italian and German. In change-of-state verbs, English exhibits no overt marking, Italian employs a reflexive-like marker (\textit{si}) in the intransitive form \citep{jezek2024semantics}, and German signals the shift through case morphology, where the patient surfaces with nominative rather than accusative case, a reflex of its syntactic subject position. 
Modern Hebrew, and other Semitic languages, show the complex non-concatenative morphology typical of this family, where verb formation relies on consonantal roots combined with vocalic templates \citep{McCarthy1979,Arad2005,Tsarfaty2004,kastner2019templatic}.

In this paper, we present curated datasets across languages for a particular task, the Blackbird Language Matrices (BLMs) \citep{merlo2023blackbirdlanguagematricesblm}, inspired by the visual analytical reasoning test of Raven Progressive Matrices (RPM, \citealt{raven1938}) and analogous to the visual ARC corpus \cite{chollet2019}. This task matches well the underling linguistic rules of alternations. BLMs are linguistic puzzles that implicitly describe a paradigmatic linguistic system. The BLM task  requires selecting the sentence that satisfies an underlying linguistic rule in an incomplete template. Specifically, the task consists in a context set of sentences that implicitly provides all necessary information for continuation of a given linguistic pattern, and an answer set of minimally differing contrastive sentences, where only one -- the missing element in the pattern described by the context -- is correct. Such template is then instantiated with curated datasets.
%The task is then performed exploring the sentence embeddings of the dataset instances, to detect how linguistic information is encoded in sentence representations \citep{nastase-merlo-2024-tracking}, in our case the relations across verb alternants. 

In this work, we explore three typologies of templates and three types of methodology for data creation and augmentation. 
%
%BLMs are valuable in many ways. We name two. They allow us to isolate which forms (e.g. a verbal alternant) within a paradigm may be subject to biases or are more easily learned by models, and, in case of errors, the contrastive error set has more diagnostic power than simple binary judgments.
%
%While the main contribution of this work lies in the description and curation of datasets from a linguistic perspective and the different types of diagnostics for the context, 
%
We also provide baseline experiments to demonstrate the diagnostic usefulness of the data and offer a  point of comparison of  model abilities. These experiments are not intended as a comprehensive evaluation. %  We establish reference results against which more sophisticated architectures can be tested, and at the same time highlighting the potential and the challenges posed by the verb alternations captured in our data.

\section{Related work}
%\input{latex/sections/related}
%\paragraph{Verb Alternations and Argument Structure.}
Verb alternations have long been a central topic in linguistic theory \citep{Levin93, rappaporthovav2024variable} and have recently attracted increasing attention in computational studies. LLMs have demonstrated strong performance on a range of alternation phenomena \citep{kann-etal-2019-verb, warstadt-etal-2019-investigating, wilson2023abstract}. For instance, \citet{yi-etal-2022-probing} show that LLMs with contextual embeddings capture information about verb alternation classes at both the word and sentence levels in English. Similarly, the semantic properties of verb arguments, such as agents and patients, have been analyzed using transformer-based models \citep{proietti-etal-2022-bert}. Beyond English, recent studies have examined verb subcategorization in Italian, assessing how well models identify and classify syntactic frames \citep{SimonettiJezekVetere2024, GrassoLoveraRulfi2024}. 

Crafting paradigm-based datasets is crucial to evaluate models and testing their ability to generalize across paradigms, for example, morphological patterns \citep{batsuren-etal-2022-unimorph, sigmorphon-2024}. Evaluating the linguistic competence of pretrained LLMs typically relies on benchmark suites combining synthetic and natural datasets. Synthetic datasets are designed to probe specific grammatical phenomena in a controlled way, often using minimal pairs or constructed paradigms \citep{warstadt-etal-2019-investigating,warstadt-etal-2020-blimp}. Recent work has explored the use of LLMs for data generation, producing synthetic corpora or targeted examples and assess their distance from natural corpus data \citep{schepens-etal-2025-llmcorpora, Civico2025}. Although automatic generation enables large-scale testing, it also raises concerns about distributional biases \citep{zhang2025doestrainingsyntheticdata, nadas-etal-2025-survey,griffiths2024bayes}. 

In contrast, natural datasets such as the Universal Dependencies treebanks (UD)  provide large-scale, annotated corpora derived from authentic textual sources \cite{de2021universal}. These treebanks have recently been used to study and create large-scale data targetting complex linguistic patterns \citep{jumelet-etal-2025-multiblimp}. UD data have also been  used to evaluate LLMs performance on tasks such as morphological analysis and syntactic annotation \citep{akkurt-etal-2024-evaluating, Acs_Hamerlik_Schwartz_Smith_Kornai_2024}.

\section{The BLM templates}

\begin{figure}[!h]
\centering
\small
\setlength{\tabcolsep}{2pt} 
\begin{tabular}{ll}
%\hline
\multicolumn{2}{c}{\sc Context}\\
\hline
1 & \textcolor{violet}{The artist} broke \textcolor{orange}{the vase} \textcolor{blue}{two years ago}\\
2 & \textcolor{violet}{The artist} broke \textcolor{orange}{the vase}  \\
3 & \textcolor{orange}{The vase} broke \textcolor{blue}{two years ago} \\
4 & ??? \\

\multicolumn{2}{c}{\sc Answer Set} \\
\hline
1 & \textcolor{orange}{The vase} broke \textcolor{violet}{the artist} \\
2 & \textbf{\textcolor{orange}{The vase} broke} \\
3 & \textcolor{violet}{The artist} broke \\
4 & \textcolor{orange}{The vase} broke \textcolor{purple}{in the museum}\\
%\hline
\end{tabular}
\caption{Simplified example of a BLM template.}
\label{fig:BLM-overview}
\end{figure}

Blackbird Language Matrices (BLMs) are linguistic puzzles designed to implicitly capture linguistic paradigms \citep{merlo2023blackbirdlanguagematricesblm,merlo2023-findings,an-etal-2023-blm,samo-etal-2023-blm,nastase2024exploring1,nastase2024exploring2,jiang-etal-2024-blm,jiang-merlo2026EACL}. In a BLM task, given an informative but incomplete context, the goal is to select the sentence that satisfies an underlying linguistic rule and completes the context. The sentence is selected from a contrastive answer set. Contexts are designed based on templates and are then instantiated using natural language instances.  A simplified BLM is shown in Figure~\ref{fig:BLM-overview}. %The instantiation of the curated dataset  \textcolor{red}{rephrase here} the task is performed on sentence embeddings to evaluate how linguistic information is represented \citep{nastase-merlo-2024-tracking}.
BLMs create a controlled environment to study how internal representations encode knowledge of linguistic paradigms, such as verb alternation.
%By analyzing continuations that follow syntactic or semantic regularities, 
To solve the puzzle, a model must observe all variants and represent the templatic structure underlying the system \citep{bobaljik2015suppletion}.

%We evaluate pretrained sentence embeddings in BLM environments where training and test sets are strictly separated. Different instantiation of the same phenomenon shows a learning space that do represent the space in common between the structures. Task performance reflects the extent to which models capture linguistic generalizations. Error analysis further reveals how models solve the task, what biases they may exhibit (e.g., an Agent bias; \citealt{huber2023surprisal}), and whether models' decisions are driven by genuine linguistic cues or by superficial character-level signals.

%In this paper, we describe the template and the data augmentation process. %, these two elements jointly define the structure of our experimental design and determine how linguistic generalizations are represented and tested within the model. 
%
%First we generate templates of BLMs and then instantiate them. Each of these steps can be accomplished in many ways and the instantiation step can use both natural data and synthetic data.

%Templates and instantiation are two key factors of a BLM matrix. 

\paragraph{Templates}
%As shown in Figure \ref{fig:BLM-overview}, the BLM consists of two core elements: a context set and an answer set. The context set consists of a series of sentences that follow a continuation pattern determined by linguistic rules. The answer set includes the correct response along with minimally differing incorrect alternatives. 

Let us first consider the context. 
In Figure \ref{fig:BLM-overview}, 
verb alternants are relatively easy to locate in a BLM template: one (or more) of the alternants is to be given in the context, and the correct answer shows the  missing alternant. The composition of the sentences in the context should highlight the underlying alternation rules. Formally, the relations operations \textit{R} transform a list of sentences into a predictable sequence \citep{merlo2023blackbirdlanguagematricesblm}.

%\textcolor{red}{AGGIUNGERE E DESCRIVERE LE RELATIONS OPERATIONS} 

These operators can be broadly divided into three categories: operators peripheral to the phenomenon under investigation (type A), operators showing relations among components of the phenomenon through linguistic diagnostics (type B), and  a combination of the two (type C).

The BLM template in Figure \ref{fig:BLM-overview} can be seen as a case of the first type (type A). The core components of the verb alternation are the verb and its arguments. The temporal constituent \textit{two years ago} is external to the linguistic phenomenon, but it helps create a pattern to be solved. In Figure \ref{fig:BLM-overview}, 
the extra-phenomenon rule is simply the presence or absence of a temporal prepositional phrase (\textit{two years ago}) between the two alternants. The correct answer is therefore \#2 (in bold).

A second type of template (type B) is designed so that models can implicitly learn internal rules by assigning properties to the relevant arguments using linguistic diagnostics drawn from grammatical descriptions \cite{samo-etal-2023-blm,jiang-merlo2026EACL}. This template allows us to test whether the model encodes the abstract linguistic properties of the phenomenon. A variation of Figure \ref{fig:BLM-overview}'s context with this type of template is shown in Figure \ref{fig:abduction-example}.
The figure shows that the two arguments of the verb are plausible subjects in a subject relative clause with an active verb, therefore both constituents can be subjects in a standard canonical clause. 

\begin{figure}[!h]
\centering
\small
\setlength{\tabcolsep}{2pt} 
\begin{tabular}{ll}
%\hline
\multicolumn{2}{c}{\sc Context}\\
\hline
1 & The artist broke the vase\\
2 & The artist that broke the vase is there \\
3 & The vase that broke is there \\
4 & ??? \\
%\hline
\end{tabular}
\caption{Example of template of type B}
\label{fig:abduction-example}
\end{figure}

The third type of template is a combination of the two. In type C (see example in Figure \ref{fig:type-III}), the context provides different kinds of cues, including linguistic diagnostics (such as relative clauses) and factors that are external to alternation (presence of prepositional phrases). The correct answer \textit{The vase broke} appears within a more complex paradigm, since the other verbs in the context display different inflectional forms (in this case, tense) and therefore require the model to rely on different relations operations.

\begin{figure}[!h]
\centering
\small
\setlength{\tabcolsep}{2pt} 
\begin{tabular}{ll}
%\hline
\multicolumn{2}{c}{\sc Context}\\
\hline

1 & The artist will break the vase on stage\\
2 & The vase will break during the performance \\
3 & We met the artist that broke the vase on stage \\
4 & The vase that broke on stage now values millions \\
5 & The artist breaks the vase \\
6 & ??? \\
%\hline
\end{tabular}
\caption{Example of template of type C}
\label{fig:type-III}
\end{figure}

The organization of a given template determines also the answer set. %The answer set consists of the correct answer and incorrect options (errors) that differ minimally from the correct answer, but designed to be informative about the type of error and its distance from the correct answer.
Answer sets contain two types of errors, that  apply to all the templates described here. 
%
%\begin{enumerate}
%    \item 
%
    \textsc{Sequence} errors, where the candidate answer does not follow the expected order of constituents in the target answer (e.g. answers 1 and 4 in Figure \ref{fig:BLM-overview}). 
%
%    \item 
\textsc{Grammar} errors, where the quality or the constituents (their semantic or syntactic nature) of the sequence does not match the target answer (e.g. answer 3 in Figure \ref{fig:BLM-overview}).
%\end{enumerate}

\paragraph{Data Augmentation}

Templates allow us to define a specific linguistic pattern and maintain full control over its structural properties. However, to ensure that the model’s results are reliable and not overly sensitive to minimal lexical instantiation, a large amount of instantiated templates is required. 
The instantiation process typically begins with one or two carefully crafted examples. The examples are then augmented in a linguistically-informed way.
In section \ref{data}, we illustrate three strategies: the use of masked modelling for controlled data generation, prompt-based augmentation, and the use of natural data extracted from large-scale corpora. 
%
%Our natural data are extracted from syntatically annotated treebanks of Universal Dependencies \citep{de2021universal} where the phenomenon under investigation -- the Hebrew verbal paradigm is morphologically annotated.
%
For each BLM matrix, two different levels of instantiations are output:  minimal lexical variation across constituents (henceforth, \textsc{MinLex}), where the template is built with the same lexical material (see Figure \ref{fig:BLM-overview} and \ref{fig:abduction-example}), and maximal lexical variation (\textsc{MaxLex}), in which sentences do not share the same lexical constituents, but they still encode the same syntactic and semantic structure, as in Figure \ref{fig:maxlex}. 

\begin{figure}[!h]
\centering
\small
\setlength{\tabcolsep}{2pt} 
\begin{tabular}{ll}
%\hline
\multicolumn{2}{c}{\sc Context}\\
\hline
1 & The queen chipped the scepter this morning\\
2 & The artist broke the vase \\
3 & The door opened at midnight \\
4 & ??? \\

\hline
\multicolumn{2}{c}{\sc Answer Set} \\
%\hline
1 & The inspector closed the case \\
2 & \textbf{The situation improved }\\
3 & The artist broke \\
4 & My roommate fried last night \\
%\hline
\end{tabular}
\caption{A simplified example of \textsc{MaxLex} of the template of Figure \ref{fig:BLM-overview}. The sentences follow the same syntax, but the words vary in each sentence.}
\label{fig:maxlex}
\end{figure}

In what follows, we present the datasets on verb alternations across languages.

\section{Data}
\label{data}
For each dataset, we introduce the linguistic phenomenon, the template, and the data augmentation procedure. Table \ref{tab:datasets-overview} provides an overview of these datasets\footnote{The datasets are available at \url{https://www.idiap.ch/en/scientific-research/data}, where
they can be found by searching with the string ‘BLM’. }.

\begin{table}[h!]
\centering
\begin{tabular}{p{1cm} p{1.5cm} p{1.3cm} p{2.1cm}}
\hline
\textbf{Name} & \textbf{Language} & \textbf{Template} & \textbf{Augmentation Strategies} \\
\hline
OD & En, It & A & Masking \\
COS    & En, It         & A      & Masking       \\
COS+   & De, It, En     & B     & Prompting     \\
Caus   & He             & C & Natural data \\
\hline
\end{tabular}
\caption{Overview of the datasets.}
\label{tab:datasets-overview}
\end{table}

\subsection{BLM-COS and BLM-OD
%: Change-of-State and Object Drop in English and Italian
}

The verbs adopted as examples in Figure \ref{fig:intro-example} belong to two distinct, well-studied verb classes \citep{Levin93}:  Change-of-State (COS) verbs (e.g., \textit{break}), and  Object-Drop (OD) verbs (e.g.,  \textit{paint}) \citep{Levin93,merlo2001automatic}. As mentioned in the introduction, these two semantic classes of verbs differ in their idiosyncratic lexical meaning, but only minimally in their argument syntactic structure properties. We restrict our investigation to English and Italian, two languages that offer useful contrasts for examining these verb classes, as Italian overtly marks the intransitive form of COS verbs but not that of OD verbs (see Figure \ref{fig:intro-example}).

\paragraph{Template}

The progression relies on superficial cues. Passive constructions are adopted as forms of intermediate stages for intransitive verbs, where the patient is the subject of the clause -- but also an initial constituent -- but the verb is inflected with a passive voice. For COS verbs, this progression is natural, while for OD verbs, the passive acts as a confounding factor in the context (cf. \citealt{MEO2007359}). 
The semantic role of the subject in the final contextual sentence is the key differentiator between the two verb classes. The context template also includes an operator peripheral to the phenomenon under investigation, the alternation between a \prep-NP and a \by-NP phrase preceded by a by-phrase (\textit{by} in English, \textit{da} in Italian).

The answer set %in the BLM task 
is constructed to reflect both the linguistic rules under study and the extra-linguistic operator. All sentences in the answer set follow a common XP–V–XP structure to favour comparison. We create eight candidates that correspond to selected combinations of four factors: agents (\textcolor{violet}{Ag}) and patients (\textcolor{orange}{Pat}), active (\textcolor{brown}{Akt}) and passive (\textcolor{teal}{Pass}), the presence or absence of a \textcolor{red}{by} preposition and the use of arguments as elements following \textcolor{red}{by} so that the correct answer requires the model to recognize the appropriate intransitive form and templatic continuation, while incorrect answers systematically deviate in terms of the semantic roles of the constituents and the voice of the verb. Errors are categorized based on their nature: if the alternation with a \textit{by}-phrase is correct but the semantics is wrong, it is classified as a \textsc{Grammar} error; if an NP appears in a transitive or passive sentence where it should not, it is classified as a \textsc{Sequence} error. Among \textsc{Sequence} errors, we emphasize that standard canonical transitive and passive clauses can appear as grammatical answers, thereby misleading the model. Figure \ref{tab:templateBLM-CosOD} shows the template for English.

For Italian, we created two additional answers due to the use of the reflexive-like element \textit{si}, characteristic of causative constructions \citep{Vietri2019}, e.g. \textit{il burro \textbf{si} scioglie} `the butter SI melts'. Its absence in COS sentences renders them ungrammatical (e.g. *\textit{il burro scioglie)}, while its presence in OD contexts produces a grammatical but transitive clause with a reflexive (\textit{l'artista si dipinge}, lit. `the artist paints herself').

\begin{figure}[!h]
\centering
\scriptsize % Adjust the font size here
\setlength{\tabcolsep}{2pt} % Adjust column separation if needed

% First two tables
\begin{tabular}{lllll} 
\hline
\multicolumn{5}{c}{\sc COS context}\\
\hline
1 & \ag & \akt & \pat & \prep-NP \\
2 & \ag & \akt & \pat  & \by-NP  \\
3 & \pat & \pass & \by-\ag & \prep-NP \\
4 & \pat & \pass & \by-\ag & \by-NP \\
5 & \pat & \pass & & \prep-NP \\
6 & \pat & \pass & & \by-NP \\
7 & \pat & \akt & & \prep-NP \\ 
8 & ??? & & &  \\ 
\end{tabular}
\hspace{0.5cm} % Adjust the space between the tables if needed
%if I use the macros everything is messed up
\begin{tabular}{ll|l} \hline
\multicolumn{3}{c}{\sc COS answers}  \\ \hline
1 & \textcolor{orange}{Pat} \textcolor{brown}{Akt}   \textcolor{red}{by}-NP & \textsc{\textbf{Correct}}\\ %0000
2 & \textcolor{violet}{Ag} \textcolor{brown}{Akt}   \textcolor{red}{by}-NP & \textsc{Grammar}  \\ %1000 
3 & \textcolor{orange}{Pat} \textcolor{teal}{Pass}  \textcolor{red}{by}-\textcolor{violet}{Ag} & \textsc{Sequence}\\
4 & \textcolor{violet}{Ag} \textcolor{teal}{Pass}   \textcolor{red}{by}-\textcolor{orange}{Pat} & \textsc{Sequence}\\
5 & \textcolor{orange}{Pat} \textcolor{brown}{Akt}   \textcolor{violet}{Ag} & \textsc{Sequence}\\
6 & \textcolor{violet}{Ag} \textcolor{brown}{Akt}   \textcolor{orange}{Pat} & \textsc{Sequence} \\
7 & \textcolor{orange}{Pat} \textcolor{brown}{Akt}   \textcolor{red}{by}-\textcolor{violet}{Ag} & \textsc{Grammar} \\
8 & \textcolor{violet}{Ag} \textcolor{brown}{Akt}   \textcolor{red}{by}-\textcolor{orange}{Pat} & \textsc{Grammar}  \\ 
\end{tabular}

% Last two tables
\vspace{0.5cm} % Adjust the vertical space if needed
\begin{tabular}{lllll} 
\hline
\multicolumn{5}{c}{\sc OD context}\\
\hline
1 & \ag & \akt & \pat & \prep-NP\\
2 & \ag & \akt & \pat & \by-NP \\
3 & \pat & \pass & \by-\ag  & \prep-NP\\
4 & \pat & \pass & \by-\ag  & \by-NP\\
5 & \pat & \pass & & \prep-NP\\
6 & \pat & \pass & & \by-NP\\
7 & \ag & \akt & & \prep-NP\\
8 & ??? & & &  \\
\end{tabular}
\hspace{0.5cm} % Adjust the space between the tables if needed
\begin{tabular}{rl|l} \hline
\multicolumn{3}{c}{\sc OD answers}  \\ \hline
1 & \textcolor{orange}{Pat} \textcolor{brown}{Akt}   \textcolor{red}{by}-NP & \textsc{Grammar}\\
2 & \textcolor{violet}{Ag} \textcolor{brown}{Akt}   \textcolor{red}{by}-NP & \textsc{\textbf{Correct}}  \\ 
3 & \textcolor{orange}{Pat} \textcolor{teal}{Pass}  \textcolor{red}{by}-\textcolor{violet}{Ag} & \textsc{Sequence}\\
4 & \textcolor{violet}{Ag} \textcolor{teal}{Pass}   \textcolor{red}{by}-\textcolor{orange}{Pat} & \textsc{Sequence}\\
5 & \textcolor{orange}{Pat} \textcolor{brown}{Akt}   \textcolor{violet}{Ag} & \textsc{Sequence}\\
6 & \textcolor{violet}{Ag} \textcolor{brown}{Akt}   \textcolor{orange}{Pat} & \textsc{Sequence} \\
7 & \textcolor{orange}{Pat} \textcolor{brown}{Akt}   \textcolor{red}{by}-\textcolor{violet}{Ag} & \textsc{Grammar} \\
8 & \textcolor{violet}{Ag} \textcolor{brown}{Akt}   \textcolor{red}{by}-\textcolor{orange}{Pat} & \textsc{Grammar}  \\ 

\end{tabular}
\caption{BLM context sets (left) for the change-of-state group (COS) and the object drop (OD) class and answers sets (right). Italian COS and OD has two additional answers in which the presence/absence of the reflexive-like element \textit{si} is absent (both GRAMMAR errors).}
\label{tab:templateBLM-CosOD}
\end{figure}

\paragraph{Data instantiation}

We selected 30 verbs in English from the classification in \citet[240-247]{Levin93} and their equivalent, in terms of meaning and syntactic behaviour, in Italian. The semi-automatic generation of the synthetic lexical data (arguments and prepositional phrases) began with the English dataset: lexical items were manually selected from the top 25 candidates produced automatically with masked language modeling. A different language model   than the one used for the sentence embeddings of our benchmark was used to minimize potential biases (\textit{bert-base-uncased}, \cite{devlin19}). Specifically, verbs were masked while only functional elements (e.g. pronominal entities) surrounded them, retrieving appropriate agents and patients. For instance, to generate arguments for the verb \textit{break}, the patient was masked in a sentence with a pronominal subject (e.g., \textit{she broke (the/a/some/...) \texttt{<MASK>}}), and the subject was masked  in a configuration with a pronominal patient to retrieve plausible agents (e.g., (the/a/some/...) \texttt{<MASK>} broke it). Figure \ref{fig:COS-MinLex} shows an example of instantiation of the BLM-COS for English (\textsc{MinLex}).

The Italian lexical elements were translations of the English data, with manual adjustments where necessary to ensure sentence acceptability, semantic plausibility, and a balanced distribution of nouns across gender and number. An example of a \textsc{MaxLex} version of an OD instantiation for Italian is shown in Figure \ref{fig:OD-Maxlex}. %For the experiment described in Section \ref{experiments}, we sampled 3,000 sentences for \textsc{MinLex}. \textsc{MaxLex} sentences were created by randomly shuffling sentences across the matrix sentences of the context and the answer set.

\begin{figure}[h!]
\centering
\small
%\scriptsize 
\begin{tabular}{l}
\hline
\textsc{Context}  \\ \hline 
The chef melted the butter on the stove \\
The chef melted the butter by mistake \\
The butter was melted by the chef on the stove \\
The butter was melted by the chef by mistake \\
The butter was melted on the stove\\
The butter was melted by mistake \\
The butter melted on the stove \\
? \\ 
\hline
\textsc{Answer Set} \\ \hline
\textbf{The butter melted by mistake} \\
The chef melted by mistake \\
The butter was melted by the chef \\
The chef was melted by the butter \\
The butter melted the chef \\
The chef melted the butter \\
The butter melted by the chef \\
The chef melted by the butter \\
\end{tabular}
\caption{An example for COS for English with minimal lexical variation (MinLEX).}
\label{fig:COS-MinLex}
\end{figure}

\begin{figure}[h!]
\centering
\scriptsize
%\small
\begin{tabular}{l}
\hline
\textsc{Context} \\ \hline
\textit{Gli artisti dipinsero questi paesaggi in una giornata} \\
\textcolor{gray}{`The artists painted these landscapes in one day'} \\
\textit{Le attrici cantavano le stesse canzoni da qualche anno} \\
\textcolor{gray}{`The actresses had been singing the same songs for a few years'} \\
\textit{Dei loghi saranno disegnati dalla ditta in meno di un'ora} \\
\textcolor{gray}{`Some logos will be drawn by the company in less than one hour'} \\
\textit{Il grano deve essere seminato dal contadino da stamattina} \\
\textcolor{gray}{`The wheat must be sown by the farmer since (lit. by) this morning'} \\
\textit{I messaggi saranno recitati con la giusta interpretazione} \\
\textcolor{gray}{`The messages will be recited with the right interpretation'} \\
\textit{Le case furono intagliate da manuale} \\
\textcolor{gray}{`The houses were carved to perfection (lit. by the handbook)'} \\
\textit{Il falegname scolpiva nella cava di marmo} \\
\textcolor{gray}{`The carpenter was carving in the marble quarry''} \\
? \\
\hline
\textsc{Answer Set} \\ \hline
\textit{Una gonna si cuciva da qualche giorno} \\
\textcolor{gray}{`A skirt SI sew for (lit. by) a few days'} \\
\textit{Le artiste si scolpiscono da qualche mese} \\
\textcolor{gray}{`The artists SI sculpt for (lit. by) a few months'} \\
\textit{I capelli possono essere lavati dall'assistente} \\
\textcolor{gray}{`The hair can be washed by the assistant'} \\
\textit{I suoi genitori devono essere cucinati dal pranzo} \\
\textcolor{gray}{`His parents must be cooked by the lunch'} \\
\textit{Le uova cucinavano le mie amiche} \\
\textcolor{gray}{`The eggs were cooking my friends'} \\
\textit{Gli studenti pescano salmoni} \\
\textcolor{gray}{`The students fish salmons'} \\
\textit{Un articolo lesse dalle autrici} \\
\textcolor{gray}{`An article read by the authors'} \\
\textit{Il sottocuoco deve impastare dalla pizza} \\
\textcolor{gray}{`The sous-chef must knead by the pizza'} \\
\textit{Gli articoli scrivono da manuale} \\
\textcolor{gray}{`The articles write to perfection (lit. by the handbook)'} \\
\textbf{\textit{Alcuni colleghi studiano da anni}} \\
\textcolor{gray}{``Some colleagues have been studying for (lit. by) years'} \\
\end{tabular}
\caption{An example for OD for Italian (with English glosses in grey) with maximal lexical variation (\textsc{MaxLex}).}
\label{fig:OD-Maxlex}
\end{figure}

\subsection{BLM-COS+: Change-of-State Food dataset}

%\paragraph{Linguistic phenomenon}

As shown in Figure \ref{fig:intro-example}, COS verb alternation can vary in its morphosyntactic strategies across languages. We built a dataset specifically for the COS verb class, but we limited our lexical domain to food-related verbs and constituents. Food-related contexts were chosen to clearly distinguish agent-like elements from patients. In choosing food-related terms, we aimed to make this dataset dynamic, easily translatable into many languages. At this stage, we have created data for English, German, and Italian.

\paragraph{Template}

The template uses relative clauses as a linguistic diagnostic. Relative clauses reveal the syntactic roles of the constituents, showing that the object of a transitive COS verb can appear either as the object in an object relative clause or as the syntactic subject in a subject relative clause in a COS intransitive configuration. The matrices can vary in the directionality of the alternants, they can start with the transitive verb in the context and probe the intransitive (T2I) or the inverse (I2T).

The answer set is minimal, since the answer only shows the combination of the arguments, their number and their position. Specifically, it includes one correct answer (the relevant alternant) and two types of errors:  one involving a reverse-role transitive (object–verb–subject), and  one involving an intransitive construction in which the agent appears as the subject. The reverse-role transitive is classified as a \textsc{Grammar} error when the target sentence is the transitive alternant, and as a \textsc{Sequence} error when the target is the intransitive alternant. The two templates (T2I and I2T) are given in Figure \ref{tab:templateBLM-T2I-I2T}.

\begin{figure}[!h]
\centering
\scriptsize
\setlength{\tabcolsep}{2pt} % adjust column separation

% T2I context and answers
\begin{tabular}{ll}
\hline
\multicolumn{2}{c}{\sc T2I Context} \\
\hline
SVO   & \textcolor{violet}{Ag} Verb \textcolor{orange}{Pa} \\
S\textsubscript{ag}R  & \textcolor{violet}{Ag} \textcolor{teal}{rel} Verb \textcolor{orange}{Pa} \\
OR  & \textcolor{orange}{Pa} \textcolor{teal}{rel} \textcolor{violet}{Ag} Verb \\
S\textsubscript{pa}R & \textcolor{orange}{Pa} \textcolor{teal}{rel} Verb \\
\hline
& Answer Set \\ \hline
\textsc{Grammar} & \textcolor{violet}{Ag} Verb \\
\textbf{\textsc{Correct}} & \textcolor{orange}{Pa} Verb \\
\textsc{Sequence} & \textcolor{orange}{Pa} Verb \textcolor{violet}{Ag} \\
\end{tabular}
\hspace{1cm}
% I2T context and answers
\begin{tabular}{ll}
\hline
\multicolumn{2}{c}{\sc I2T Context} \\
\hline
SVO   & \textcolor{orange}{Pa} Verb \\ 
S\textsubscript{pa}R & \textcolor{orange}{Pa} \textcolor{teal}{rel} Verb \\
OR    & \textcolor{orange}{Pa} \textcolor{teal}{rel} \textcolor{violet}{Ag} Verb \\
S\textsubscript{ag}R  & \textcolor{violet}{Ag} \textcolor{teal}{rel} Verb \textcolor{orange}{Pa} \\
\hline

& Answer Set \\ \hline
\textsc{Sequence} & \textcolor{violet}{Ag} Verb \\
\textbf{\textsc{Correct}} & \textcolor{violet}{Ag} Verb \textcolor{orange}{Pa}\\
\textsc{Grammar} & \textcolor{orange}{Pa} Verb \textcolor{violet}{Ag} \\
\end{tabular}

\caption{BLM templates for the T2I and I2T groups. Context sets  and corresponding Answer sets are shown with colour-coded elements: agent (Ag), patient (Pa), verb (V), and relative markers (rel).}
\label{tab:templateBLM-T2I-I2T}
\end{figure}

\paragraph{Data instantiation}

We selected 25 verbs for each language that show the very same structure. For example, for Italian we excluded the verb \textit{bollire} `boil' since the intransitive form does not require the \textit{si} \citep{Vietri2019} and for German, we did not include separable verbs (e.g. \textit{aufwärmen} 'reheat') to have comparable results across the morphological patterns. 

We augmented the data using the DeepSeek-V3 model \citep{guo2025deepseekr1}, adopting a hybrid approach that combines LLM generation with explicit linguistic constraints. This ensured the generated sentences remained grammatically well-formed and semantically coherent within each alternation paradigm. For each target language, the prompt requested the generation of 80 possible patients for each verb (25 interactions) and a total of 300 possible agents across verbs (1 interaction). To assess the quality of the augmented data, native speakers evaluated the combinations of the seed constituents.

As illustrated in Figure \ref{fig:intro-example}, German exhibits case marking that differentiates patients in object and subject positions. This marking is overtly distinct only for masculine nouns, but not for feminine or neuter constituents. Therefore, we created two datasets for German: one containing only masculine, case-marked constituents (Case or C) and one with masculine, feminine and neuter arguments (Mix).

\begin{figure}[h!]
\centering
\small
\begin{tabular}{p{7cm}}
\hline
\textsc{Context}  \\ \hline 
\textit{Der Chef schmiltzt den Käse} \\
\textcolor{gray}{`The.NOM chef melts the.ACC cheese'} \\
\textit{Der Chef der den Käse schmiltzt} \\
\textcolor{gray}{`The.NOM chef that.NOM  the.ACC  cheese melts'} \\
\textit{Den Käse den der Chef schmiltzt} \\
\textcolor{gray}{`The.ACC cheese that.ACC  the.NOM chef melts'} \\
\textit{Der Käse der schmiltzt} \\
\textcolor{gray}{`The.NOM that.NOM cheese melts'} \\
\hline
\textsc{Answer Set} \\ \hline
\textit{Der Chef schmiltzt} \\
\textcolor{gray}{`The.NOM chef melts'} \\
\textit{Der Käse schmiltzt} \\
\textcolor{gray}{`The.NOM cheese melts'} \\
\textit{Den Käse schmiltzt der Chef} \\
\textcolor{gray}{`The.NOM cheese is melted by the.ACC chef'} \\
\end{tabular}
\caption{MinLex for COS+T2IC (Case) for German. English glosses in grey. NOM = Nominative case, ACC = Accusative case.}
\label{tab:templateBLM-T2I-MaxLex}

\end{figure}

\subsection{BLM-CausH: Hebrew binyanim system}
\label{data:heb}

\begin{figure}
    \centering
    \includegraphics[width=1\linewidth]{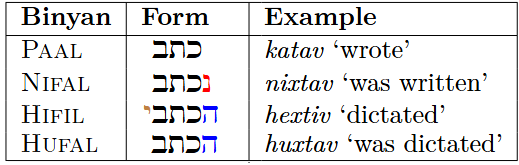}
    \caption{Binyanim under investigations and relative examples for the root KTB (related to the act of writing). Adapted from \citealt[574-575] {kastner2019templatic}. The colour codes the different overt characters representing morphemes marking the alternation.}
    \label{fig:ex-hebrew}
\end{figure}

Hebrew verbs exhibit distinct morphological patterns that encode voice and causativity called \textit{binyanim}. In our investigation we select to work on four forms \citep{samo2026modellingmorphologyverbalparadigms}, presented in Figure \ref{fig:ex-hebrew}, that encode different types of marking: \textsc{Paal} represents the basic, labile transitive form; \textsc{Nifal} typically marks the passive voice \citep[71]{coffin2005reference}; \textsc{Hifil} and \textsc{Hufal} correspond to causative active and passive forms.

\paragraph{Template}

The context set contains three complete pairs of sentences and one incomplete pair. In the first three pairs, verbs are inflected according to a specific binyan. In the fourth pair, one sentence illustrates the remaining binyan, and the task is to infer the missing sentence. The answer set comprises four sentences, each exemplifying a different binyan . This is to be considered a type C template, since the model will rely on the forms but also on the number of arguments, their semantic nature and their context. %Except for the last pair, the order of the context pairs is rotated across all possible sequences for each potential correct answer.

\begin{figure}[!h]
\centering
\footnotesize
\setlength{\tabcolsep}{2pt} % adjust column separation

\begin{tabular}{lllllll}
\hline
\multicolumn{7}{c}{\sc Context} \\
\hline
 \textcolor{red}{Paal} &  & \textcolor{red}{Paal} & & \textcolor{red}{Paal} & & \textcolor{teal}{Nifal}\\
 \textcolor{red}{Paal}  &  & \textcolor{red}{Paal}  & & \textcolor{red}{Paal} & & \textcolor{teal}{Nifal}\\
 \textcolor{teal}{Nifal} &  & \textcolor{teal}{Nifal} & &  \textcolor{orange}{Hifil} & & \textcolor{orange}{Hifil}\\
 \textcolor{teal}{Nifal}  & ... & \textcolor{teal}{Nifal} & ... &  \textcolor{orange}{Hifil} & ... & \textcolor{orange}{Hifil}\\
 \textcolor{orange}{Hifil} &  & \textcolor{blue}{Hufal} & & \textcolor{blue}{Hufal} & & \textcolor{blue}{Hufal} \\
 \textcolor{orange}{Hifil}  &  & \textcolor{blue}{Hufal} & & \textcolor{blue}{Hufal} & & \textcolor{blue}{Hufal} \\
 \textcolor{blue}{Hufal}  & & \textcolor{orange}{Hifil} & &  \textcolor{teal}{Nifal} & & \textcolor{red}{Paal} \\
 \multicolumn{7}{c}{\textbf{?}}  \\
\hline
\multicolumn{7}{c}{\sc Answer Set}\\ \hline
\multicolumn{7}{c}{\textcolor{red}{Paal}} \\
\multicolumn{7}{c}{\textcolor{teal}{Nifal}}  \\
\multicolumn{7}{c}{\textcolor{orange}{Hifil}}  \\
\multicolumn{7}{c}{\textcolor{blue}{Hufal}} \\
\end{tabular}

\caption{BLM template for the CausH.}
\label{tab:templateBLM-T2I}
\end{figure}

\paragraph{Data instatiation}

We create two BLM datasets. One dataset, \textsc{Natural}, was created with natural occurring sentences extracted from UD treebanks. The second dataset, \textsc{Synthetic}, only contains the verb corresponding to each sentence in the \textsc{Sentence} dataset, with no additional lexical content. These sentences can be grammatically correct since Hebrew allows both subject and object drop \citep{vainikka1999empty,erteschik2013missing}. The data were retrieved using \textit{match.grew.fr} \citep{guillaume-2021-graph}, by querying a simple variable X with the relevant  annotation in two treebanks of Hebrew containing respectively news (HBT, \citealt{tsarfaty2013unified,mcdonald2013universal}; 114,648 tokens, 6,143 trees) and encyclopaedic entries (IAHLTWiki, henceforth IW; \citealt{ZeldesHowellOrdanBenMoshe2022}; 103,395 tokens; 5,039 trees). Both treebanks are version 2.15. The query collects sentences where the main verb is annotated with relevant binyan for the morphosyntactic property \textsc{HebBinyan}\footnote{E.g. \textsc{Paal}: \textit{pattern} X [HebBinyan = "PAAL"]}.

Both datasets only instantiate the BLM templates with the most complex form of lexical variation (\textsc{MaxLex}). The lack of \textsc{MinLex} is because not all roots are compatible with all templatic structures \citep{kastner2019templatic}.
%, since roots share a broad semantic field and the relationships among forms within a paradigm are not always transparent (e.g., the root P\d{K}D ‘ordering/depositing’). 
All the wrong candidates are errors of \textsc{Grammar}.

\begin{figure*}
\footnotesize
    \centering
    \includegraphics[width=0.7\linewidth]{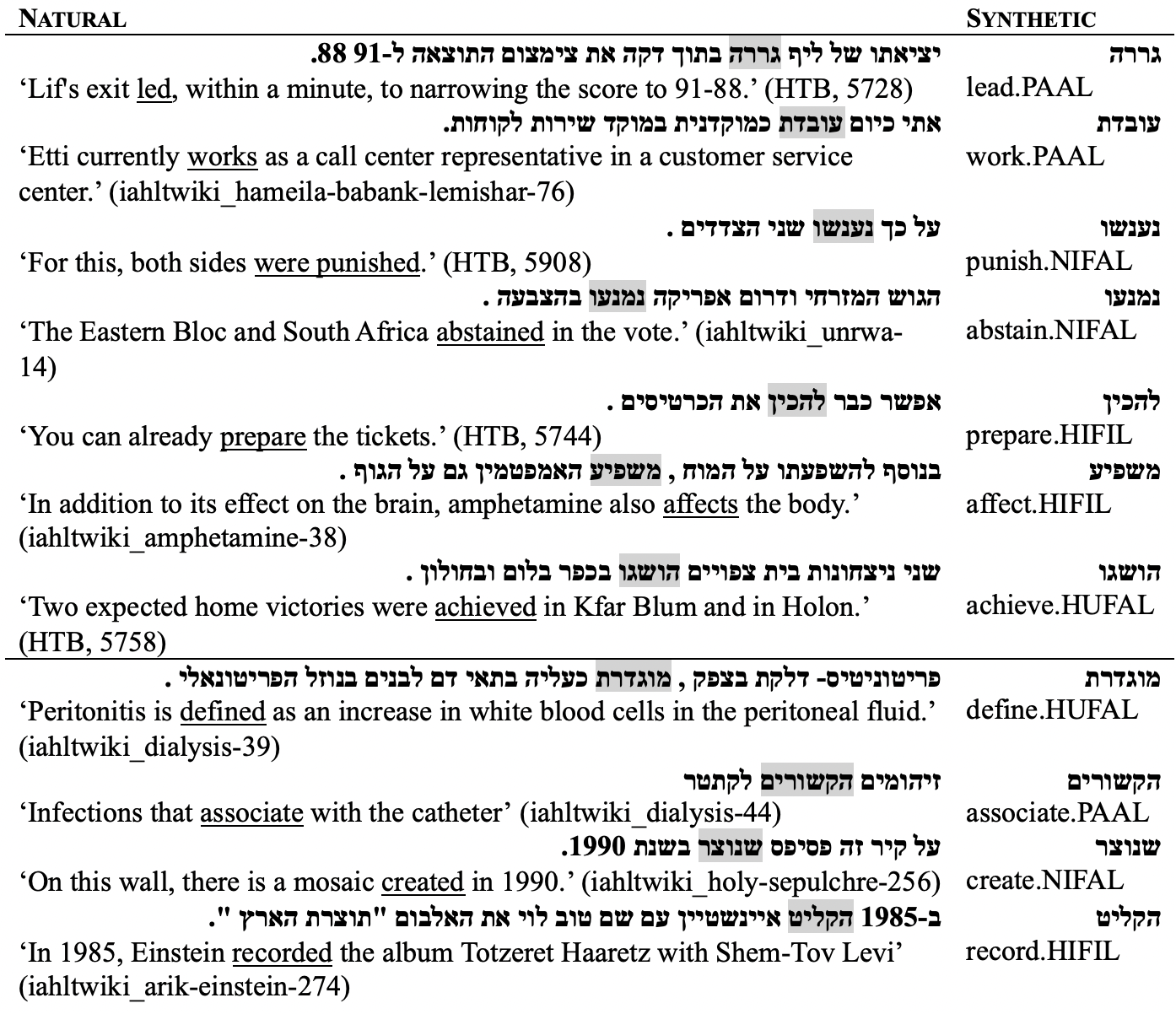}
    \caption{An example of instantiation of sentences (\textsc{Natural}) extracted from UD and the derived synthetic data made by the verbs (\textsc{Synthetic}). Glosses are provided and the verb of the sentence is highlighted in gray. }
    \label{fig:placeholder}
\end{figure*}

\section{Experiments}
\label{experiments}

%\subsection{Models}
Building on prior studies \citep{nastase-merlo-2024-tracking}, we primarily apply a Feed-Forward Neural Network (FFNN) to establish a benchmark across the system and various models. Specifically, we use a FFNN architecture as a baseline system, using sentence embeddings derived from the curated datasets as described in \citep{samo-etal-2023-blm}. For each sentence in the BLM, we compute the averaged token embeddings (the token embeddings are calculated with the multilingual model Electra (\textit{google/electra-base-discriminator}). The FFNN receives the stacked embeddings that represent the context, is trained with a max-margin loss, and selects the answer whose embedding has the highest cosine similarity to the network’s output. For Hebrew, we also compare the results of a monolingual model (AlephBERT, \textit{onlplab/alephbert-base}). Each dataset is then spit into training and testing. To avoid data leakage, we use disjoint sets of training and testing instances. Statistics on the dataset are given in Table \ref{tab:stats-exp}.

\begin{table}[h!]
\footnotesize
    \centering
    \begin{tabular}{lllrr}
         Data & Lang & Lex & Train & Test \\ \hline
         COS & En, It & \textsc{Min/Max} & 2700 & 300 \\
         OD & En, It & \textsc{Min/Max} & 2700 & 300 \\
         COS+\textsubscript{T2I} & De\textsubscript{Case}  &  \textsc{Min/Max} & 1600 & 400 \\
          & De\textsubscript{mix}  & \\
          & En, It & \\
          COS+\textsubscript{I2T}& De\textsubscript{Case} & \textsc{Min/Max} & 1600 & 400 \\
          & De\textsubscript{Mix} \\
          & En, It & \\
        Caus & He\textsubscript{Nat}& \textsc{Max} & 7200 & 800 \\ 
        & He\textsubscript{synth} \\ \hline
          \end{tabular}
    \caption{Datasets, languages, training and testing instances for each language and lexical variation.}
    \label{tab:stats-exp}
\end{table}

%\subsection{Results}

\begin{figure}[h!]
    \centering    \includegraphics[width=1\linewidth]{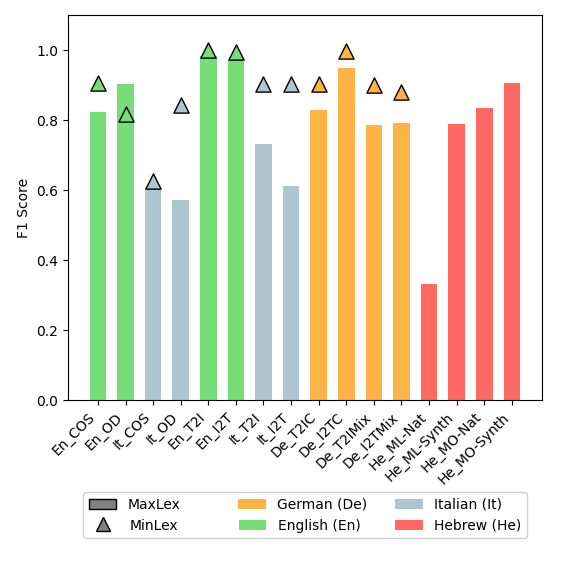}
    \caption{F1 scores across datasets.}
    \label{fig:figure_F1}
\end{figure}

Performance in terms of F1 scores is visualized in Figure \ref{fig:figure_F1}. 
English consistently achieves the highest  scores across most tasks, with higher marks in Template B (En-COS+\textsubscript{T2I}: 0.983, En-COS+\textsubscript{I2T}: 0.971). Italian demonstrates performance scores generally lower than English. For template B, Italian shows higher performance in \textsc{MaxLex} when the target answer is the marked intransitive (It-COS+\textsubscript{T2I}: 0.733). German shows higher performance in the dataset when the patient in COS verbs is case-marked differently in the two alternants (DeEN-COS+\textsubscript{IT2C}: 0.950). In Hebrew the monolingual (He\_MO) model outperform the multilingual (He\_ML) in the natural settings, but multilingual and monolingual are comparable in the synthetic setting. Finally, we can observe that \textsc{MinLex} is easier than \textsc{MaxLex} across languages, except for OD English.

\begin{figure}
    \centering
    \includegraphics[width=1\linewidth]{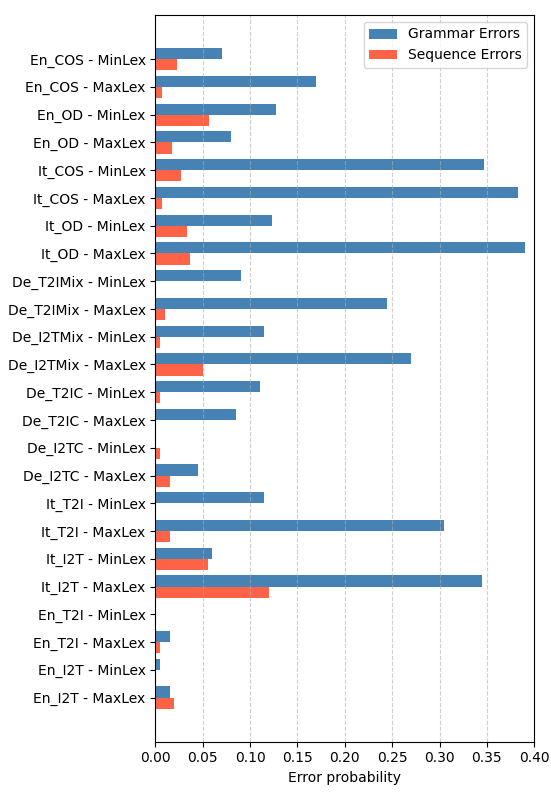}
    \caption{Error analysis across datasets.}
    \label{fig:error-analysis}
\end{figure}

The error analysis is presented in Figure~\ref{fig:error-analysis}. Overall, the FFNN learns the \textsc{Sequence} pattern of the BLM, but, with  many grammatical errors.

%original text:
%On the other hand, we remark many grammatical errors, especially those involving syntax–semantic mapping mismatches (e.g., \textit{The chef melted}). In English and Italian, the constructions exhibit relatively low grammar error rates, with slightly higher error rates in \textsc{MaxLex} conditions. For German, the case-marked constructions show minimal errors in I2T conditions. 

COS and OD primarily exhibit grammatical errors (see Table \ref{Tab:errorCOSOD-Appendix} in the Appendix). The most prominent error in both languages involves syntax–semantic mapping mismatches, where the subject of an intransitive verb is interpreted as an agent rather than as a patient (e.g., \textit{The artist broke} instead of \textit{The vase melts}). In English OD MaxLex, the same probability for this error (4\%) is assigned to the candidate answer in which a nominal phrase introduced by the preposition \textit{by} is incorrectly interpreted as agentive (e.g., \textit{The vase breaks by the artist}). In English OD MinLex, a common \textsc{Sequence} error involves a transitive sentence with an agentive subject (e.g., \textit{The artist paints the vase}). In Italian COS, we also observe that the absence of the reflexive-like element \textit{si} (\textit{il burro stava sciogliendo da qualche minuto}) or its presence in OD (\textit{Le artiste si scolpiscono da qualche mese}) is not a prominent error.

With respect to COS+, we observe that, for German, BLMs comprising only case-marked constructions perform better than mixed ones, especially in I2T conditions. English shows very few errors, while Italian does not show differences between the T2I and I2T configurations. Also in this case the errors are mainly error of syntax-semantic interface, (e.g., \textit{The chef melted} instead of \textit{The butter melts}).

\begin{figure}[h!]
    \centering    \includegraphics[width=1\linewidth]{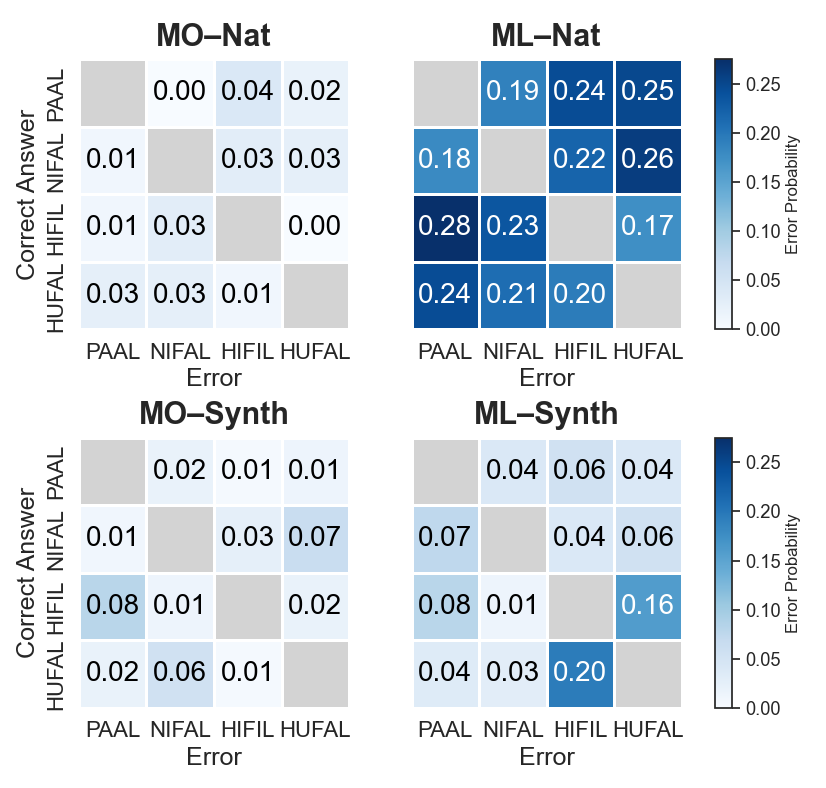}
    \caption{Confusion matrices for Hebrew across datasets and models, for each verbal voice, reported as proportions.}
    \label{fig:figure_HebrewEA}
\end{figure}

Figure \ref{fig:figure_HebrewEA} shows a confusion matrix in terms of error probability of Hebrew. All the errors in the Hebrew data are grammatical. In the monolingual both datasets, we do not observe a preferred answer. On the other hand, the multilingual natural data shows an error pattern in line with a chance level performance and in the multilingual synthetic data there is clear confusion between \textsc{Hifil} and \textsc{Hufal}, two verbal voices that share common morphological marking.

\section{Discussion}
\label{dicussion}
BLM curated datasets allow  us to probe systematic grammatical knowledge in models by providing controlled, contrastive templates that isolate the phenomena under investigation.
Templates play a crucial role in our approach: they are designed both to isolate the linguistic phenomenon under investigation and to instruct the model in how to generalize from controlled input. 

The templates presented here show different types of complexity.
%
%\begin{enumerate}
%\item 
The BLM template used for the dataset BLM-COS and BLM-OD (type A) requires the model to pay attention to an extra-linguistic phenomenon  to find the sequence, and in so doing it helps identify the linguistic problem at stake. This level can be seen as \textit{analogical} in a weak sense: the model learns to associate surface patterns without necessarily accessing the internal structure of the target phenomenon.

%\item 
The BLM template for the dataset BLM-COS+ (Type B) forces analogical representations in a stronger sense: the model must map relational structures rather than surface similarities. This step introduces a form of \textit{abductive} inference, since the model must hypothesize the most plausible underlying relation that accounts for the observed patterns.

%\item 
The BLM template adopted for the dataset BLM-CausH focuses on the relation across elements within a paradigm. This third type targets \textit{systematicity} — the capacity to represent a grammatical paradigm. Systematicity implies that once a model acquires a relation within a system, it should be able to extend that relation to other members of the same paradigm in a consistent manner. %We can claim that this template does require forms of analogy and abduction.

Finally, results from our baseline experiment suggest that multilingual models achieve comparable cross-linguistic performance with synthetic data. %\cite{Nastase_CLIC-IT2025_2025}. 
Comparing monolingual, as well as other generative models, will provide additional avenues for investigation in the understanding of linguistic knowledge in models.

\section{Conclusions}
\label{conclusions}
In this work, we presented datasets in several languages withing the framework of the Blackbird Language Matrices (BLMs), designed to probe linguistic knowledge in LLMs for verb alternations. By combining carefully curated templates with linguistically informed data augmentation strategies, we provide datasets for testing cross-sentence syntactic and semantic generalizations.

Overall, our work demonstrates the importance of linguistically-principled dataset design and provides resources for cross-linguistic evaluation. It also establishes a foundation for future research investigating the limits of model generalization across paradigms and languages. %Future work should expand the range of language families and linguistic phenomena examined.

\section*{Limitations}

%Future work could address the limitations of this contribution by expanding language coverage, exploring additional models, and performing comprehensive validation. \textcolor{red}{To be changed}

Despite the controlled design and cross-linguistic scope of the proposed datasets, several limitations should be acknowledged. The datasets focus on a restricted set of verb alternation phenomena and languages. While English, Italian, German, and Hebrew provide typologically diverse case studies, they do not fully represent the range of morphosyntactic strategies found cross-linguistically.

The experimental evaluation is intentionally limited to baseline architectures and embedding-based representations. The reported experiments are designed to demonstrate the diagnostic usefulness of the datasets and the task rather than to provide a comprehensive comparison across state-of-the-art LLMs. Consequently, the results should not be interpreted as definitive statements about current model capabilities.

%\textcolor{red}{Future work could address the limitations of this contribution by expanding language coverage, exploring additional models, and performing comprehensive validation.}

\section*{Ethics}
A portion of the data in the datasets used in this work are derived from publicly available corpora, including sources such as news articles and other open-access materials. These datasets may contain content that is sensitive or potentially distressing. We recognize that some individuals may find certain topics upsetting.

\section*{Acknowledgments}
We gratefully acknowledge the support of this work by the Swiss National Science Foundation, through grant SNF Advanced grant TMAG-1\_209426 to PM. 

\section{Bibliographical References}\label{sec:reference}

\bibliographystyle{lrec2026-natbib}
\bibliography{latex/custom,latex/references}

\label{lr:ref}
\bibliographystylelanguageresource{lrec2026-natbib}
\bibliographylanguageresource{languageresource}

\clearpage

\appendix
\onecolumn
\section*{Appendix}

\begin{table}[h!]
\centering
\begin{tabular}{llrrrr}
\multicolumn{6}{c}{ENGLISH} \\ \hline
Wrong Answer    & Error Type & COS\textsc{MinLex} & COS\textsc{MaxLex} & OD\textsc{MinLex} & OD\textsc{MaxLex} \\ \hline 
1  \textcolor{orange}{Pat} \textcolor{brown}{Akt} \textcolor{red}{by}-NP    & \textsc{Grammar}    & -           & -           & 35         & 12         \\
2  \textcolor{violet}{Ag} \textcolor{brown}{Akt} \textcolor{red}{by}-NP    & \textsc{Grammar}    & 21          & 49          & -          & -          \\
3  \textcolor{orange}{Pat} \textcolor{teal}{Pass} \textcolor{red}{by}-\textcolor{violet}{Ag}   & \textsc{Sequence}   & 1           & 1           & 0          & 0          \\
4 \textcolor{violet}{Ag} \textcolor{teal}{Pass} \textcolor{red}{by}-\textcolor{orange}{Pat}   & \textsc{Sequence}     & 4           & 0           & 1          & 2          \\
5  \textcolor{orange}{Pat} \textcolor{brown}{Akt} \textcolor{violet}{Ag}       & \textsc{Sequence}     & 2           & 1           & 3          & 0          \\
6 \textcolor{violet}{Ag} \textcolor{brown}{Akt} \textcolor{orange}{Pat}     & \textsc{Sequence}     & 0           & 0           & 13         & 3          \\
7 \textcolor{orange}{Pat} \textcolor{brown}{Akt}  \textcolor{red}{by}-\textcolor{orange}{Pat}     & \textsc{Grammar}    & 0           & 2           & 0          & 0          \\
8 \textcolor{violet}{Ag} \textcolor{brown}{Akt}  \textcolor{red}{by}-\textcolor{orange}{Pat}    & \textsc{Grammar}    & 0           & 0           & 3          & 12         \\
\\
\multicolumn{6}{c}{\textsc{Italian}} \\ \hline
Wrong Answer    & Error Type & COS\textsc{MinLex} & COS\textsc{MaxLex} & OD\textsc{MinLex} & OD\textsc{MaxLex} \\ \hline 
1  \textcolor{orange}{Pat} SI \textcolor{brown}{Akt} \textcolor{red}{by}-NP & \textsc{Grammar}     & -           & -           & 0          & 7          \\
2 \textcolor{violet}{Ag} SI \textcolor{brown}{Akt} \textcolor{red}{by}-NP  & \textsc{Grammar}     & 103         & 109         & 5          & 15         \\
3  \textcolor{orange}{Pat} \textcolor{teal}{Pass} \textcolor{red}{by}-\textcolor{violet}{Ag}   & \textsc{Sequence}     & 1           & 7           & 0          & 0          \\
4 \textcolor{violet}{Ag} \textcolor{teal}{Pass} \textcolor{red}{by}-\textcolor{orange}{Pat}   & \textsc{Sequence}     & 7           & 2           & 1          & 0          \\
5  \textcolor{orange}{Pat} \textcolor{brown}{Akt} \textcolor{violet}{Ag}       & \textsc{Sequence}     & 0           & 0           & 4          & 9          \\
6 \textcolor{violet}{Ag}  \textcolor{brown}{Akt}  \textcolor{orange}{Pat}    & \textsc{Sequence}     & 0           & 0           & 5          & 2          \\
7  \textcolor{orange}{Pat} \textcolor{brown}{Akt} \textcolor{red}{by}-\textcolor{violet}{Ag}     & \textsc{Grammar}     & 0           & 4           & 5          & 9          \\
8 \textcolor{violet}{Ag}  \textcolor{brown}{Akt} \textcolor{red}{by}- \textcolor{orange}{Pat}    & \textsc{Grammar}     & 0           & 1           & 5          & 15         \\
9  \textcolor{orange}{Pat} \textcolor{brown}{Akt} \textcolor{red}{by}-NP     & \textsc{Grammar}     & 26          & 76          & 1          & 0          \\
10 \textcolor{violet}{Ag}  \textcolor{brown}{Akt} \textcolor{red}{by}-NP     & \textsc{Grammar}     & 21          & 49          & -          & -         
\end{tabular}
\caption{Raw counts of errors for BLM-COS and OD in English and Italian across lexical variation conditions (test set \textit{n} = 300).}
\label{Tab:errorCOSOD-Appendix}
\end{table}

\end{document}